\pdfoutput=1
\documentclass[11pt]{article}

\usepackage{acl}
\usepackage{times}
\usepackage{latexsym}

\usepackage[T1]{fontenc}
\usepackage[utf8]{inputenc}

\usepackage{microtype}

\usepackage{inconsolata}
\usepackage{graphicx} 
\usepackage{amsmath}
\usepackage{booktabs}
\usepackage{multirow} 
\usepackage{amssymb}
\usepackage[most]{tcolorbox}
\usepackage{listings}
\usepackage{cleveref}
\usepackage{pifont}
\usepackage{amssymb}

\usepackage{enumitem}
\usepackage{algorithm}
\usepackage{algpseudocode}

\algrenewcommand\algorithmicindent{1em}

\algrenewcommand\algorithmicfor{\textbf{for}}
\algrenewcommand\algorithmicend{\textbf{end}}
\algrenewcommand\algorithmicdo{\textbf{do}}
\algrenewcommand\algorithmicif{\textbf{if}}
\algrenewcommand\algorithmicthen{\textbf{then}}
\algrenewcommand\algorithmicreturn{\textbf{return}}

\makeatletter
\renewcommand{\ALG@beginalgorithmic}{\small}
\makeatother

\algrenewcommand\alglinenumber[1]{}

\usepackage{xcolor}  % for color options
\usepackage{todonotes}

\usepackage{subcaption}

\lstdefinelanguage{json}{
  basicstyle=\ttfamily,
  morestring=[b]",
  morecomment=[l]{//},
  morekeywords={true,false,null},
  sensitive=true,
  stringstyle=\color{orange},
  commentstyle=\color{gray},
  keywordstyle=\color{blue},
  literate=
    *{:}{{{\color{black}:}}}{1}
     {,}{{{\color{black},}}}{1}
     {[}{{{\color{purple}[}}}{1}
     {]}{{{\color{purple}]}}}{1}
}

\makeatletter
\def\@makefnmark{\hbox{\@textsuperscript{\normalfont\@thefnmark}}}
\makeatother

\title{Xiangqi-R1: Enhancing Spatial Strategic Reasoning in LLMs \\ for Chinese Chess via Reinforcement Learning}

\author{
  Yuhao Chen, Shuochen Liu, Yuanjie Lyu, Chao Zhang, Jiayao Shi, Tong Xu\thanks{ Corresponding author.} \\
  University of Science and Technology of China \\
  \texttt{isyuhaochen@mail.ustc.edu.cn} \\
  \texttt{tongxu@ustc.edu.cn}
}

\begin{document}
\maketitle

\begin{abstract}
Game playing has long served as a fundamental benchmark for evaluating Artificial General Intelligence. While Large Language Models (LLMs) have demonstrated impressive capabilities in general reasoning, their effectiveness in spatial strategic reasoning, which is critical for complex and fully observable board games, remains insufficiently explored. In this work, we adopt Chinese Chess (Xiangqi) as a challenging and rich testbed due to its intricate rules and spatial complexity. To advance LLMs’ strategic competence in such environments, we propose a training framework tailored to Xiangqi, built upon a large-scale dataset of five million board-move pairs enhanced with expert annotations and engine evaluations. Building on this foundation, we introduce \textbf{Xiangqi-R1}, a 7B-parameter model trained in multi-stage manner. Our Experimental results indicate that, despite their size and power, general-purpose LLMs struggle to achieve satisfactory performance in these tasks. Compared to general-purpose LLMs, Xiangqi-R1 greatly advances with an 18\% rise in move legality and a 22\% boost in analysis accuracy. Our results point to a promising path for creating general strategic intelligence in complex areas.
\footnote{Code: 
\href{https://github.com/isyuhaochen/Xiangqi-R1}{https://github.com/isyuhaochen/Xiangqi-R1}.
}

%: (1) fine-tuning for legal move prediction to capture basic spatial rules, (2) incorporating strategic annotations to improve decision-making, and (3) applying reinforcement learning via Group Relative Policy Optimization (GRPO) with multi-dimensional reward signals to enhance reasoning stability.

\end{abstract}

\section{Introduction}
\begin{figure}[t]
\centering
\includegraphics[width=\linewidth]{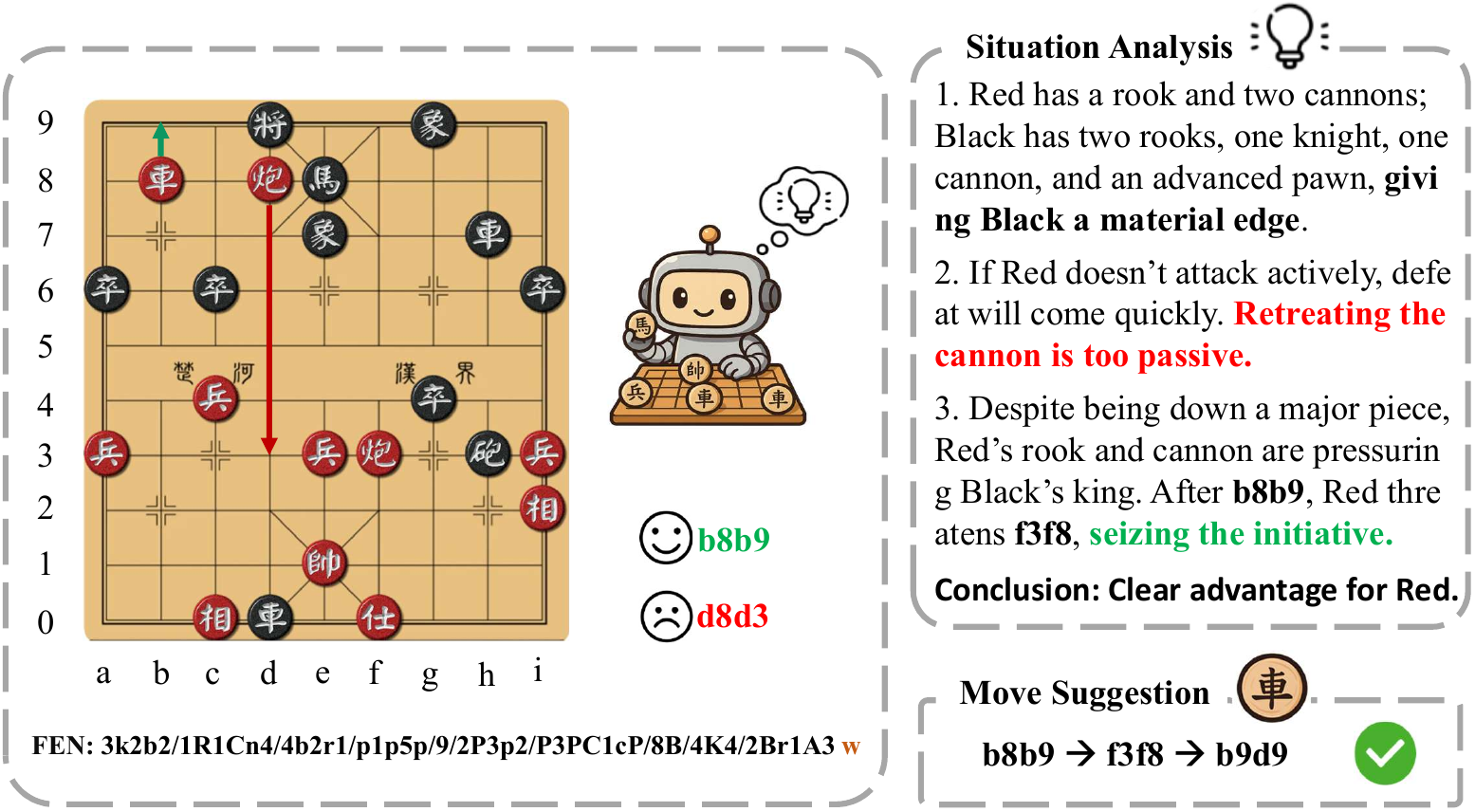} % Reduce the figure size so that it is slightly narrower than the column.
\caption{The Spatial Strategic Reasoning capability involves generating a coherent situation analysis and an appropriate move suggestion, conditioned on a given board state.}
\label{fig1}
\end{figure}

Game playing has long served as a key benchmark for evaluating intelligence, particularly in the pursuit of Artificial General Intelligence (AGI). In recent years, Large Language Models (LLMs) have demonstrated remarkable general reasoning capabilities, achieving performance comparable to or even exceeding that of human experts across a wide range of tasks~\cite{guo2025deepseek, yang2025qwen3}. However, their capabilities in fully observable strategic game tasks involving spatial rule-based reasoning and strategic decision-making remain inadequately validated. This gap constitutes a critical challenge for LLMs' progression towards more generalized AI. Such tasks require models to both understand complex rules and make strategically sound, legal moves based on the current board. We collectively refer to these comprehensive abilities as Spatial Strategic Reasoning. This term refers to the model's ability to accurately interpret the structure of a 2D board state, perform effective analysis, and generate actions based solely on text input, as illustrated in Fig.~\ref{fig1}.
% todo
% 现有工作是用sft来提高模型的能力，在国际象棋中有一些的进展。
% 所以有两个局限，第一个局限是使用sft是否足够？，第二个局限是在国际象棋上评估是否足够具有挑战性？

Efforts to overcome this challenge have focused on exploring supervised fine-tuning (SFT) methods for improving move generation in games. Notable progress has been made in the domain of chess with pioneering works such as ChessGPT~\cite{feng2023chessgpt}, ChessLLM~\cite{zhang2025complete}, and MATE~\cite{wang2024explore}. Despite this progress, these studies still have two key limitations related to spatial strategic reasoning. One limitation lies at the modeling level, and the other lies at the evaluation level. 
\textbf{First}, prior work primarily focuses on chess as the evaluation domain. While useful, its regular geometry may allow models to generate legal moves without a deep understanding of spatial constraints. 
We argue that Chinese Chess (Xiangqi) serves as a more challenging and informative testbed. Compared to chess, Xiangqi has a larger board (9x10) and more types of pieces (7), and it imposes stricter spatial constraints (e.g., Nine Palaces). These factors require a deeper understanding of board structure and spatial reasoning.
\textbf{Second}, SFT improves recall and structural coherence of generated outputs~\cite{chu2025sft}. However, SFT models trained on static game datasets tend to memorize training patterns rather than develop genuine strategic understanding, which leads to struggles with novel board states and poor identification of optimal moves in unfamiliar situations.

To address these research gaps, we introduce a comprehensive dataset for Xiangqi and propose a reasoning model tailored for this domain, named \textbf{Xiangqi-R1}. For dataset construction, we collected professional game records from open-source Xiangqi platforms to build a large-scale dataset comprising five million board–move pairs. In addition, we curated tens of thousands of high-quality annotated samples, including move commentaries written by human experts and advantage evaluations as well as optimal move annotations generated by the Pikafish engine~\cite{pikafish2025}, providing rich semantic and strategic information to support advanced reasoning tasks.

To effectively utilize this data, especially the semantically rich and complex commentary content (as shown in Fig.~\ref{fig1}), we propose a multi-stage training framework to enhance LLMs' game reasoning capabilities: (1) Fine-tuning based on large-scale legal move data to enhance the model's understanding of spatial rules; (2) Integrating expert commentary and engine-annotated data to improve the model's board state assessment and strategic explanation capabilities; (3) Introducing a reinforcement learning (RL) mechanism using a small amount of data with multi-dimensional rewards (Move Reward, Analysis Reward, and Format Reward) to enhance the model's reasoning stability in complex games through policy optimization.

Experimental results validate the effectiveness of our approach. Despite its modest scale of 7B parameters, Xiangqi-R1 significantly outperforms general-purpose models several times larger across multiple key metrics in spatial strategic reasoning. For instance, compared to the strong model Doubao-Pro-1.5-Think~\cite{seed2025seed1}, the move legality metric (legal@1) improved by \textbf{18\%}, and the board evaluation accuracy (3-class@1) improved by \textbf{22\%}, demonstrating that our method significantly enhanced the model's spatial strategic reasoning capabilities in this scenario. In summary, this research makes three key contributions:

\noindent\textbf{1}) We introduce the first Xiangqi-specific evaluation framework that comprehensively assesses LLMs’ spatial strategic reasoning across different game stages and piece types, enabling fine-grained performance analysis.

\noindent\textbf{2}) We propose a training strategy that integrates expert knowledge, engine annotations, and a multi-dimensional RL reward mechanism to enhance the model’s spatial strategic reasoning capabilities.

\noindent\textbf{3}) Experimental results show that despite having only 7B parameters, Xiangqi-R1 surpasses larger general LLMs in Xiangqi spatial reasoning, exhibiting superior reasoning and strategic adaptability.

\section{Related Work}
\subsection{Artificial Intelligence in Chess-like Games}
The development of AI has reshaped traditional board games such as chess, turning them into testbeds for strategic reasoning and platforms for algorithmic innovation. From Deep Blue’s brute-force search and handcrafted evaluation functions to Stockfish’s $\alpha$–$\beta$ pruning and refined heuristics, the field has shifted from raw computation to more sophisticated modeling~\cite{campbell2002deep, stockfish2025}.  Inspired by AlphaGo~\cite{silver2016mastering} and AlphaZero~\cite{silver2017mastering}, the chess community integrated neural network evaluation modules (e.g., NNUE), giving rise to hybrid engines like Pikafish~\cite{pikafish2025} that combine deep feature extraction with classical search, improving decision-making in complex positions. As chess engines now surpass human capabilities, research has shifted from human competition to using game tasks to assess planning, causal reasoning, and structural generalization in general-purpose AI.

Recent studies have begun exploring the integration of LLMs into chess to examine their potential in sequential decision-making and structural understanding. Systems such as ChessGPT~\cite{feng2023chessgpt}, ChessLLM~\cite{zhang2025complete}, and MATE~\cite{wang2024explore} incorporate expert moves, engine evaluations, and tactical annotations through SFT, showing preliminary success in strategic move generation and evaluation. Building on this, we extend LLMs' architectures to Xiangqi, addressing its unique spatial constraints and asymmetric rules. Our method shows improved comprehension and adaptability in complex, rule-intensive environments, contributing to research on general strategic intelligence.
%guo2025deepseek,openai2025competitiveprogramminglargereasoning,jin2025search

% zhang2025tearagtokenefficientagenticretrievalaugmented, chen2025think，zhang2025aisalesmanreliablelargelanguage, lin2025speculativedecodingreimaginedmultimodal, liu2025lookthinkunifyingreasoning
\subsection{Reinforcement Learning for LLMs Reasoning}
RL has demonstrated significant potential in enhancing the reasoning capabilities of LLMs across various domains such as mathematics, code generation~\cite{zhang2025tearagtokenefficientagenticretrievalaugmented, chen2025think, zhang2025aisalesmanreliablelargelanguage, lin2025speculativedecodingreimaginedmultimodal, liu2025lookthinkunifyingreasoning}. These tasks often require complex, multi-step decision-making processes, which pose substantial challenges for traditional SFT. Initially, RL from Human Feedback (RLHF)~\cite{christiano2017deep,ouyang2022training} was introduced to align model outputs with human preferences, addressing the limitations of purely supervised approaches. To tackle the complexity of actor-critic methods like Proximal Policy Optimization (PPO)~\cite{schulman2017proximal}, more efficient alternatives have emerged. Direct Preference Optimization (DPO)~\cite{rafailov2023direct} simplifies training by removing the critic component, though its off-policy nature constrains generalization. 

More recently, GRPO~\cite{shao2024deepseekmath} has improved training stability through enhanced advantage estimation techniques. Despite these advances, RL-based methods remain underexplored in spatial strategy reasoning tasks, particularly in complex domains such as chess and Xiangqi. In this work, we extend GRPO by incorporating an expert-guided reward function, aiming to leverage RL to overcome the limitations inherent in prior SFT-only approaches and to advance research in this critical yet underdeveloped area.

\section{Method}
\begin{figure*}[t]
\centering
\includegraphics[width=\linewidth]{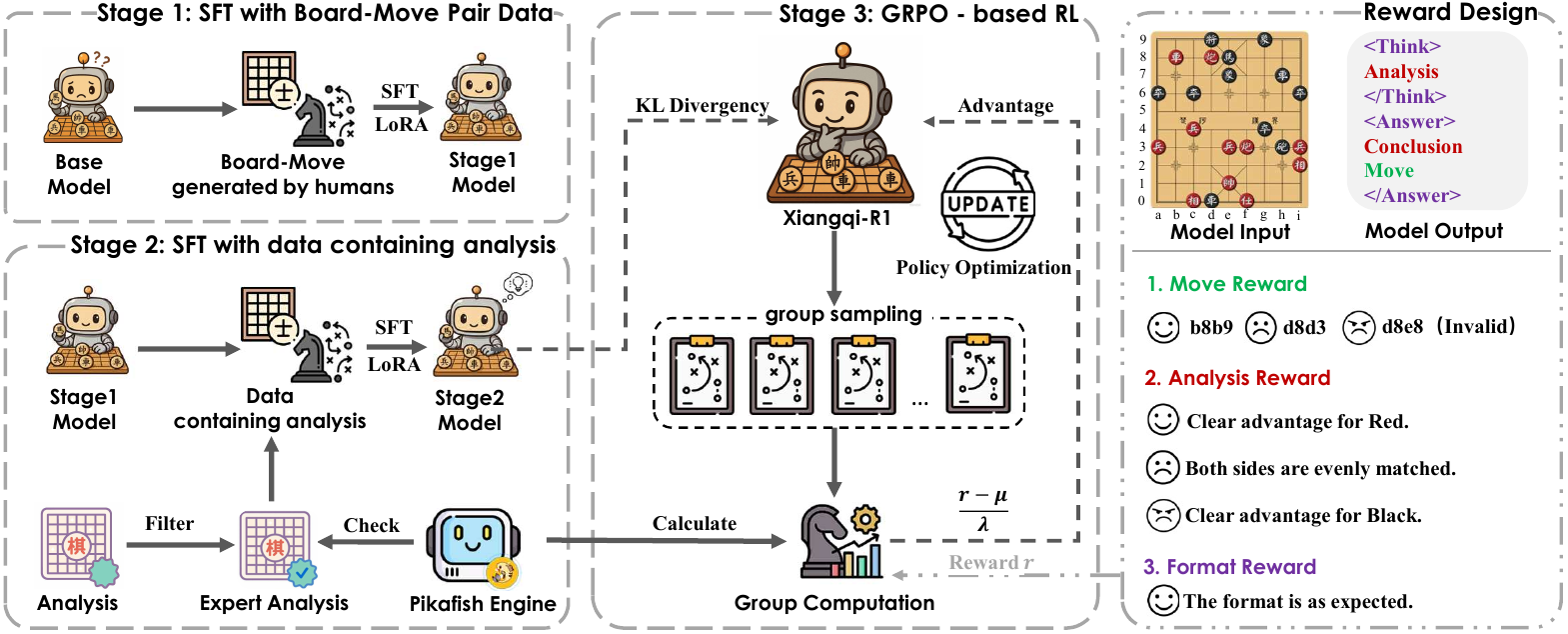} 
\caption{Overview of our proposed framework. Stage 1: SFT with Board-Move Pair Data. Stage 2: SFT with data containing analysis. Stage 3: A fine-grained reward function is constructed and combined with a GRPO-based RL method to further enhance the model’s spatial strategic reasoning capabilities.}
\label{fig2}
\end{figure*}
In this section, we provide a detailed introduction to our proposed training framework. An overview of our framework is illustrated in Fig.\ref{fig2}. 

\subsection{Preliminary}
Prior to presenting our method, we provide an overview of Xiangqi, board and move representation, the Pikafish engine, and necessary definitions to facilitate comprehension.
% todo 修改
% \paragraph{Xiangqi.} Xiangqi is a traditional two-player, fully observable, deterministic, zero-sum game played on a 9×10 grid divided by a central river. Each player controls 16 of the 32 total pieces, across seven types: \textit{King}, \textit{Rook}, \textit{Knight}, \textit{Cannon}, \textit{Elephant}, \textit{Guard}, and \textit{Pawn}. The goal is to checkmate the opponent’s \textit{King}, who is confined to a 3×3 palace at the rear center of each side and cannot face the opposing \textit{King} directly along a file without intervening pieces. Each piece has unique movement rules. Cannons move like Rooks but capture only by jumping over exactly one piece on the same rank or file. Knights move one step orthogonally, then one step diagonally outward, but can’t move if the orthogonal step is blocked. Elephants move two points diagonally, can’t cross the river, and can’t leap over pieces. Guards move one point diagonally within the palace. Pawns move one point forward, and after crossing the river, can also move sideways. The river and palaces create asymmetry and add strategic depth to both offense and defense.

\paragraph{Xiangqi.} Xiangqi is a traditional two-player, fully observable, zero-sum game played on a 9×10 grid divided by a central \textit{River}. The \textit{River} divides the board and restricts the movement of certain pieces. Each side has a 3×3 \textit{Palace} at the rear center, confining the \textit{King} and \textit{Guards}. There are seven piece types, each with unique movement rules. For a comprehensive description of the piece movement rules, see Appendix~\ref{app:xiangqi}.

\paragraph{Board Representation.}
The simplest way to represent Xiangqi positions is using a string called BoardStr, which is intuitive and human-readable. Portable Game Notation (PGN) records the entire move sequence, making it better for computer processing but less focused on the current position. Forsyth-Edwards Notation (FEN) efficiently describes the entire board state and the side to move in a single line, as shown in Fig.\ref{fig1}. In our experiments, both BoardStr and FEN are used as model inputs to improve position understanding.
\paragraph{Move Representation.}
Moves describe a piece’s transition from one position to another. The Chinese File Format (CFF) is intuitive for humans, using four Chinese characters, but is not ideal for computers. The Internet Chinese Chess Server (ICCS) format represents moves by start and end coordinates, making it more suitable for computer processing, and was one of the first formats used in computer systems.

% \paragraph{Pikafish.} It is an open-source, high-performance Xiangqi engine designed to support research and development. It offers fast search and evaluation and is compatible with standard Xiangqi protocols. Given a search depth or time and a position in FEN format, it returns the best move and the position’s evaluation: 

\paragraph{Pikafish.} Pikafish is an open-source high-performance Xiangqi engine with an Elo rating of 3954, which is far above the top human level of around 2800, and it is used as our ground truth reference. It supports fast search and evaluation, complies with standard Xiangqi protocols, and, given a search depth or time and a position in FEN format, returns the best move and position evaluation:
\begin{equation}
\text{BestMove}, \text{Value} = \text{Pikafish}(\text{Cfg}, \text{FEN}).
\end{equation}

For convenience, evaluation scores are positive for Red and negative for Black. In all experiments, we set the search depth in \text{Cfg} to 25, meaning Pikafish searches 25 plies.

\subsection{Xiangqi Dataset Construction}
To effectively train and evaluate our models, it is crucial to construct a high-quality and reliable normalized dataset. We curated a large-scale corpus of publicly available, open-source Xiangqi games from online platforms, consisting of over hundreds of thousands of matches in PGN format. Each record includes comprehensive metadata and detailed move sequences from both players, with a subset containing expert commentary. The raw PGN data were parsed and transformed into FEN using the open-source cchess library. This process yielded a structured dataset in the form of FEN–Move–Comment (optional) triplets, where each instance represents a specific game state, the corresponding move, and, when applicable, expert annotations. This pipeline systematically eliminates incomplete, low-quality entries, resulting in a curated dataset that is both representative and robust for downstream learning tasks.

Specifically, we retain only moves made by the winning side and moves from both sides in drawn games, excluding moves by the losing side. Although winners’ moves may contain errors, these are usually corrected by subsequent moves, leading to victory. In contrast, errors by the losing side tend to be critical and directly result in defeat. This filtering ensures data validity and quality. For commentary data, we filter out comments unrelated to the game and retain only those associated with high-quality moves, defined as moves whose evaluation differs from the best move by less than a threshold \(\sigma_{\text{good}}\). In all experiments, \(\sigma_{\text{good}}\) is set to 100, representing a narrow margin in the evaluation system. To ensure data privacy, we use DeepSeek-V3~\cite{deepseekai2024deepseekv3technicalreport} to remove comments containing sensitive information such as player names. The selection criteria are as follows:
\begin{equation}
\_,\ \text{NewValue} = \text{Pikafish}(\text{Cfg}, \text{FEN} + \text{Move}),
\end{equation}
\begin{equation}
|\text{NewValue} - \text{BestValue}| \leq \sigma_{\text{good}}.
\label{eq:good}
\end{equation}

After data preprocessing, we divide the dataset into three training stages and a separate test set. Stage 1 includes 5.1 million board–move pairs, while Stages 2 and 3 contain 52.9K and 52.8K annotated samples with human commentary, respectively. The test set consists of 2,800 high-quality samples spanning different phases of expert-level games. For the Stage 3 and test set, we enumerate all possible legal moves under the given FEN and compute the evaluation score for each move under the same settings to facilitate assessment.

To facilitate supervised learning, we categorize game positions based on evaluation scores provided by the Pikafish engine. Specifically, we define a five-class labeling scheme as follows:
\begin{equation}
\resizebox{\linewidth}{!}{$
\text{Situation} = 
\begin{cases}
\text{Balanced}, & |\text{Value}| \leq \sigma_s \\
\text{Slight Adv. (R/B)}, & \sigma_s < |\text{Value}| < \sigma_l \\
\text{Clear Adv. (R/B)}, & |\text{Value}| \geq \sigma_l
\end{cases}.
$}
\label{eq:situation_analysis}
\end{equation}

Here, \(\text{Value}\) denotes the evaluation score, where positive values favor Red and negative values favor Black. Based on the game analysis conventions of Pikafish, we set $\sigma_{\text{s}}=100$ and $\sigma_{\text{l}}=800$ to distinguish between balanced, slightly advantageous, and significantly advantageous positions. To improve robustness and reduce sensitivity to threshold tuning, we also introduce a simplified three-class setting by merging the slight and clear advantage categories into a single ``Advantage'' class.

\subsection{Stage 1\&2: SFT on Board-Move and Analytical Data}
To equip the model with preliminary reasoning capabilities, we fine-tune it using the collected dataset. However, only approximately 1\% of the data contains human-written commentary, which limits direct supervision for strategic understanding. To effectively leverage this limited annotated subset, we adopt a two-stage fine-tuning strategy. In the first stage, we perform SFT on board–move pairs, enabling the model to interpret board states in textual format and generate reasonable move suggestions. Building upon this, the second stage further fine-tunes the model using the commentary-annotated data, allowing it to develop an initial ability to analyze and assess game situations. This hierarchical training procedure helps to preliminarily enhance the model’s spatial strategic reasoning in complex game scenarios.

\begin{table*}[t]
  \centering
  \small  % 使用10pt字体，如果还不够可以换成 \footnotesize
  \resizebox{\textwidth}{!}{
  \begin{tabular}{l|cc|cc|cc|cc|cc}
    \toprule
    \multirow{2}[4]{*}{\textbf{Models}} & \multicolumn{6}{c|}{\textbf{Move Suggestion}} & \multicolumn{4}{c}{\textbf{Situation Analysis}} \\
    \cmidrule{2-7} \cmidrule{8-11}
    &legal@1 & legal@3 & good@1 & good@3 & best@1 & best@3 & 3-class@1 & 3-class@3 & 5-class@1 & 5-class@3 \\
    \midrule
    & \multicolumn{10}{c}{\textbf{Direct Answer Model}} \\
    Qwen-2.5-7B-Instruct & 0.0121 & 0.0468 & 0.0043 & 0.0104 & 0.0007 & 0.0011 & 0.0032 & 0.0125 & 0.0011 & 0.0050 \\
    Doubao-Lite & 0.0432 & 0.0939 & 0.0111 & 0.0279 & 0.0018 & 0.0046 & 0.0246 & 0.0554 & 0.0236 & 0.0521 \\
    DeepSeek-V3 & 0.3021 & 0.4504 & 0.0864 & 0.1464 & 0.0175 & 0.0339 & 0.1225 & 0.1871 & 0.0275 & 0.0525 \\
    Doubao-Pro-1.5 & 0.3521 & 0.5350 & 0.1046 & 0.1918 & 0.0279 & 0.0529 & 0.2014 & 0.3446 & 0.1375 & 0.2432 \\
    \midrule
    & \multicolumn{10}{c}{\textbf{Reasoning Model}} \\
    R1-Distill-Qwen-7B & 0.0129 & 0.0225 & 0.0039 & 0.0086 & 0.0000 & 0.0007 & 0.0082 & 0.0150 & 0.0061 & 0.0121 \\
    DeepSeek-R1 & 0.5982 & 0.8943 & 0.1825 & 0.3650 & 0.0664 & 0.1671 & 0.2043 & 0.3736 & 0.1054 & 0.2221 \\
    Doubao-Pro-1.5-Think & 0.7721 & 0.9354 & 0.2282 & 0.3829 & 0.0829 & 0.1636 & 0.4039 & \underline{0.6696} & 0.2486 & 0.4729 \\
    \midrule
    & \multicolumn{10}{c}{\textbf{Supervised Fine-Tuning Model}}\\
    SFT-Stage1-7B & \underline{0.9221} & 0.9361 & \underline{0.4486} & 0.4721 & \textbf{0.1936} & 0.2086 & - & - & - & - \\
    SFT-Stage2-7B & 0.8643 & \underline{0.9693} & 0.3525 & \underline{0.5257} & 0.1189 & \underline{0.2321} & \underline{0.4311} & 0.6539 & \underline{0.3886} & \underline{0.5914} \\
    \midrule
    % \rowcolor{gray!15}
    Xiangqi-R1-7B (Ours) & \textbf{0.9521} & \textbf{0.9721} & \textbf{0.4607} & \textbf{0.5375} & \underline{0.1814} & \textbf{0.2489} & \textbf{0.6286} & \textbf{0.6868} & \textbf{0.6104} & \textbf{0.6646} \\
    \bottomrule
  \end{tabular}
  }
  \caption{Performance comparison of various models on the Xiangqi dataset. The table reports results on two subtasks: Move Suggestion and Situation Analysis. Metrics for Move include legal@k, good@k, and best@k, while Analysis is evaluated using 3-class@k and 5-class@k accuracy. Models are grouped into Direct Answer Models and Reasoning Models. The best results are highlighted in bold, and the second-best are underlined. Our proposed Xiangqi-R1-7B achieves state-of-the-art performance across most metrics.}
  \label{tab:result}
\end{table*}

\subsection{Stage 3: Expert-Guided GRPO-Based RL and Reward Design}
To further enhance the model's spatial strategic reasoning and generalization capabilities, we decompose this complex ability into three targeted reward objectives: \( R_{\mathrm{move}} \), encouraging accurate move generation; \( R_{\mathrm{analysis}} \), promoting effective positional analysis; and \( R_{\mathrm{format}} \), enforcing consistency and stability in the model's reasoning format. This modular reward design enables more interpretable learning signals and facilitates targeted improvements in each sub-skill.
\paragraph{Move Reward ($R_{\mathrm{move}}$).}
To enhance the model's move generation capability, we introduce a multi-level move reward function $R_{\mathrm{move}} $ that evaluates the quality of the generated move at three levels: legal, good, and best. Specifically, we define the following indicator variables to capture these criteria: 
$r_{\text{best}} = \mathbb{I}(\text{Move} = \text{BestMove})$,
$r_{\text{good}} = \mathbb{I}(\text{Move} = \text{GoodMove})$,
$r_{\text{legal}} = \mathbb{I}(\text{Move is valid})$. All judgments are made using the professional engine Pikafish.
Thus, the final reward for move generation is computed as:
\begin{equation}
R_{\mathrm{move}} = r_{\text{legal}} + r_{\text{good}} + r_{\text{best}}.
\end{equation}

The definition of GoodMove is derived according to Eq.~\ref{eq:good}. Such a reward function assigns higher rewards to the model for producing higher-quality move suggestions, while invalid moves receive a reward of zero. It is worth noting that we pre-assigned scores to all possible moves in the training positions using Pikafish in advance, so that the model can evaluate them promptly without being affected by Pikafish's thinking time.

\paragraph{Analysis Reward ($R_{\mathrm{analysis}}$).} To promote accurate and insightful strategic analysis of the board situation, we introduce the analysis reward $R_{\mathrm{analysis}}$, which evaluates whether the model's analysis of the current position aligns with the assessment provided by the high-quality engine Pikafish:
\begin{equation}
R_{\text{analysis}} = 
\begin{cases}
1, & \text{if the analysis is correct} \\
0, & \text{otherwise}
\end{cases}.
\label{eq:analysis}
\end{equation}

This reward adopts a binary scoring scheme, where the score is determined based on whether the model correctly performs the situation analysis task: a correct judgment receives a reward of 1, and 0 otherwise.

\paragraph{Format Reward ($R_{\mathrm{format}}$).} 
The model is prompted to produce both a strategic analysis and a final answer. Specifically, the analysis must be fully enclosed within \texttt{<Think>...</Think>} tags, while the final answer must appear entirely within \texttt{<Answer>...</Answer>} tags. The answer is expected to include a summary of the board situation analysis and a move suggestion. Simultaneously, if no additional content appears outside the \texttt{<Think>} and \texttt{<Answer>} blocks, such an output is considered well-formed and conforms to the expected reasoning format.
\begin{equation}
R_{\text{format}} = 
\begin{cases}
1, & \text{if the format is as expected} \\
0, & \text{otherwise}
\end{cases}.
\label{eq:format}
\end{equation}

When the model outputs an incorrect format, the rewards \( R_{\mathrm{move}} \) and \( R_{\mathrm{analysis}} \) cannot be properly assigned, resulting in a reward of zero. This mechanism encourages the model to converge more rapidly toward the desired output format.

\paragraph{Training Algorithm.} As illustrated in Fig.~\ref{fig2}, we initialize our policy with the Stage~2 model and adopt the GRPO algorithm for training. This process is guided by a carefully designed reward function as described above. Specifically, GRPO samples a group of candidate outputs for each state and computes the group-relative advantage to optimize the policy. Meanwhile, the Stage~2 model serves as a reference to calculate the Kullback-Leibler divergence between the updated policy and the reference, acting as a regularization term to prevent the policy from deviating excessively from the original distribution. This training framework effectively balances relative performance improvements among candidates and overall consistency with the reference model, ensuring both effectiveness and stability during optimization.

\section{Experiments}
\subsection{Experimental configurations}
\paragraph{Dataset.}
The Xiangqi dataset is organized into three training stages and one test phase. Stage 1 adopts a board-move format with 5.1M samples. Stage 2 uses a board-analysis-move format with 52.9K samples. Stage 3 employs a board-move-value format, where value scores are computed for all legal moves, totaling 52.8K samples. The test set consists of 2.8K samples. Details of the dataset construction are described in a preceding section.

\paragraph{Baseline.} As shown in Tab.~\ref{tab:result}, our experiments evaluate several state-of-the-art LLMs, particularly those with strong support for Chinese (with BoardStr represented in Chinese). We categorize these models into two groups: Direct Answer Models and Reasoning Models. The evaluated models include the latest versions from the Qwen~\cite{qwen2.5}, DeepSeek~\cite{guo2025deepseek,deepseekai2024deepseekv3technicalreport}, and Doubao~\cite{seed2025seed1} series. Our model, Xiangqi-R1-7B, is fine-tuned based on Qwen-2.5-7B-Instruct.
% 可以更详细一点，如果页数不够的话
\paragraph{Hyperparameters.}
All experiments were conducted on 4 * A6000 GPUs. In the training phase, we employed DeepSpeed ZeRO-2~\cite{rasley2020deepspeed, rajbhandari2021zero} and LoRA~\cite{hu2022lora} to mitigate GPU memory consumption. We use the TRL library to implement training with the GRPO algorithm. Details on the configurations can be found in the Appendix A.

\subsection{Evaluation Metrics}
\begin{table*}[t]
  \centering
  \small  % 使用10pt字体，如果还不够可以换成 \footnotesize
  \resizebox{\textwidth}{!}{
  \begin{tabular}{l|cc|cc|cc|cc|cc}
    \toprule
    \multirow{2}[4]{*}{\textbf{Models}} & \multicolumn{6}{c|}{\textbf{Move Suggestion}} & \multicolumn{4}{c}{\textbf{Situation Analysis}} \\
    \cmidrule{2-7} \cmidrule{8-11}
    & legal@1 & legal@3 & good@1 & good@3 & best@1 & best@3 & 3-class@1 & 3-class@3 & 5-class@1 & 5-class@3 \\
    \midrule
    Qwen-2.5-7B-Instruct & 0.0121 & 0.0468 & 0.0043 & 0.0104 & 0.0007 & 0.0011 & 0.0032 & 0.0125 & 0.0011 & 0.0050 \\
    SFT-Stage2-7B & 0.8643 & 0.9693 & 0.3525 & 0.5257 & 0.1189 & 0.2321 & 0.4311 & 0.6539 & 0.3886 & 0.5914 \\
    \midrule
    Xiangqi-R1-7B (Ours) & \textbf{0.9521} & \underline{0.9721} & \textbf{0.4607} & \underline{0.5375} & \textbf{0.1814} & 0.2489 & \textbf{0.6286} & \underline{0.6868} & \textbf{0.6104} & \textbf{0.6646} \\

     w/o. $R_\mathrm{move}$ & 0.8736 & 0.9657 & 0.3654 & 0.5300 & 0.1296 & 0.2364 & \underline{0.5721} & 0.6643 & \underline{0.5575} & \underline{0.6457} \\
     w/o. $R_\mathrm{analysis}$ & \underline{0.9464} & \textbf{0.9761} & \underline{0.4375} & \textbf{0.5443} & \underline{0.1664} & \textbf{0.2575} & 0.4846 & \textbf{0.6911} & 0.4446 & 0.6318 \\
     w/o. $R_\mathrm{move}, R_\mathrm{analysis}$ &0.8914 & 0.9700 & 0.3644 & 0.5346 & 0.1318 & \underline{0.2550} & 0.4457 & 0.6607 & 0.4043 & 0.5979 \\
    \bottomrule
  \end{tabular}
  }
  \caption{Performance comparison on move suggestion and situation analysis metrics. The ablation results demonstrate the impact of removing specific reward components.}
  \label{tab:ablation}
\end{table*}
% 补充一下评测说明
We evaluate the performance of models on two tasks: \textbf{Move Suggestion} and \textbf{Situation Analysis}. For the Move Suggestion task, we adopt three sampling-based metrics to quantify the model's effectiveness. Specifically, \textit{legal@k} measures the probability that at least one valid move is generated within $k$ samples. \textit{good@k} assesses whether at least one high-quality move, as detailed in Eq.~\ref{eq:good}. \textit{best@k} captures the likelihood that the model produces the best move within $k$ attempts. For the Situation Analysis task, we employ classification-based metrics to evaluate the model’s ability to assess game situations. The \textit{3-class@k} metric categorizes each situation into one of three strategic outcomes: advantage to Red, balanced, or advantage to Black. To enable a more fine-grained analysis, we further introduce the \textit{5-class@k} metric, which distinguishes between five categories: significant advantage to Red, slight advantage to Red, balanced, slight advantage to Black, and significant advantage to Black. Notably, we evaluate Situation Analysis only when the model produces a valid move, since an invalid move shows a lack of basic game understanding, making any correct assessment likely accidental.

\subsection{Main Results}
\paragraph{Despite strong performance on many NLP tasks, mainstream LLMs show clear limitations in Xiangqi.} Tab.~\ref{tab:result} provides a comprehensive evaluation of representative LLMs, spanning from lightweight 7B to ultra-large 671B parameter models, on these subtasks. Our results show a clear trend: Direct Answer Models that generate responses without explicit reasoning perform poorly across all scales. For example, Qwen-2.5-7B-Instruct reaches only 0.012 in legal@1, and even larger models such as DeepSeek-V3 achieve only 0.302. These low scores show that such models struggle with basic rule-compliant reasoning. In contrast, Reasoning Models outperform Direct Answer models on both subtasks, particularly on legal@k, reflecting better legal move generation. However, they still have difficulty producing high-quality moves, as indicated by lower good@k and best@k scores. The situation analysis task is even more challenging, and both model types perform poorly, with accuracy far below expectations. 

By contrast, most human players familiar with the rules achieve nearly 100\% valid moves. This large gap highlights a significant limitation in current LLMs’ spatial strategic reasoning. The results suggest that merely scaling model size is insufficient; integrating domain knowledge and structured reasoning is essential to improve performance in complex, rule-heavy environments.

\paragraph{Our methods significantly enhance the model's spatial strategic reasoning capabilities, including move suggestion and situation analysis.} As shown in Tab.~\ref{tab:result}, our method significantly boosts a 7B model to outperform many mainstream models, including those up to 671B parameters. In the SFT-Stage1 phase, by fine-tuning on 5 million samples, the model achieves a legal@1 score of 0.922, indicating that it can reliably generate valid, rule-compliant moves. In SFT-Stage2, we further increase the task complexity by requiring the model to provide both move suggestions and situation analysis simultaneously. This introduces a new challenge and slightly reduces move suggestion accuracy, but the model improves in situation analysis, showing enhanced reasoning under more complex objectives.

Building on this, Xiangqi-R1-7B, integrates expert-guided RL, achieving substantial improvements on both subtasks. Notably, situation analysis performance shows a significant boost, confirming the effectiveness of our approach. These results demonstrate that a relatively small 7B model with domain-specific training and iterative self-play can outperform much larger models. Our findings suggest that while general-purpose LLMs struggle with complex spatial reasoning tasks like Xiangqi, specialized training and structured learning can unlock their potential. Applying this approach to larger models may yield even greater gains.

\subsection{Ablation study}
% \begin{figure*}[t]
%     \centering
%     \begin{minipage}[b]{0.32\textwidth}
%         \includegraphics[width=\textwidth]{figs/move_full.pdf}
%         \caption{Performance variation of Xiangqi-R1 across different numbers of pieces on the board.}
%         \label{fig:move}
%     \end{minipage}
%     \hfill
%     \begin{minipage}[b]{0.32\textwidth}
%         \includegraphics[width=\textwidth]{figs/move_diff.pdf}
%         \caption{Differences in performance between Xiangqi-R1 and SFT-Stage2 across different board piece counts.}
%         \label{fig:diff}
%     \end{minipage}
%     \hfill
%     \begin{minipage}[b]{0.32\textwidth}
%         \includegraphics[width=\textwidth]{figs/pieces.pdf}
%         \caption{Performance of Xiangqi-R1 on Different Piece Types in Terms of Usage and Accuracy.}
%         \label{fig:temp}
%     \end{minipage}
% \end{figure*}
\begin{figure*}[t]
    \centering

    % --- 图1 ---
    \begin{minipage}[b]{0.3\textwidth}
        \centering
        \includegraphics[width=\textwidth]{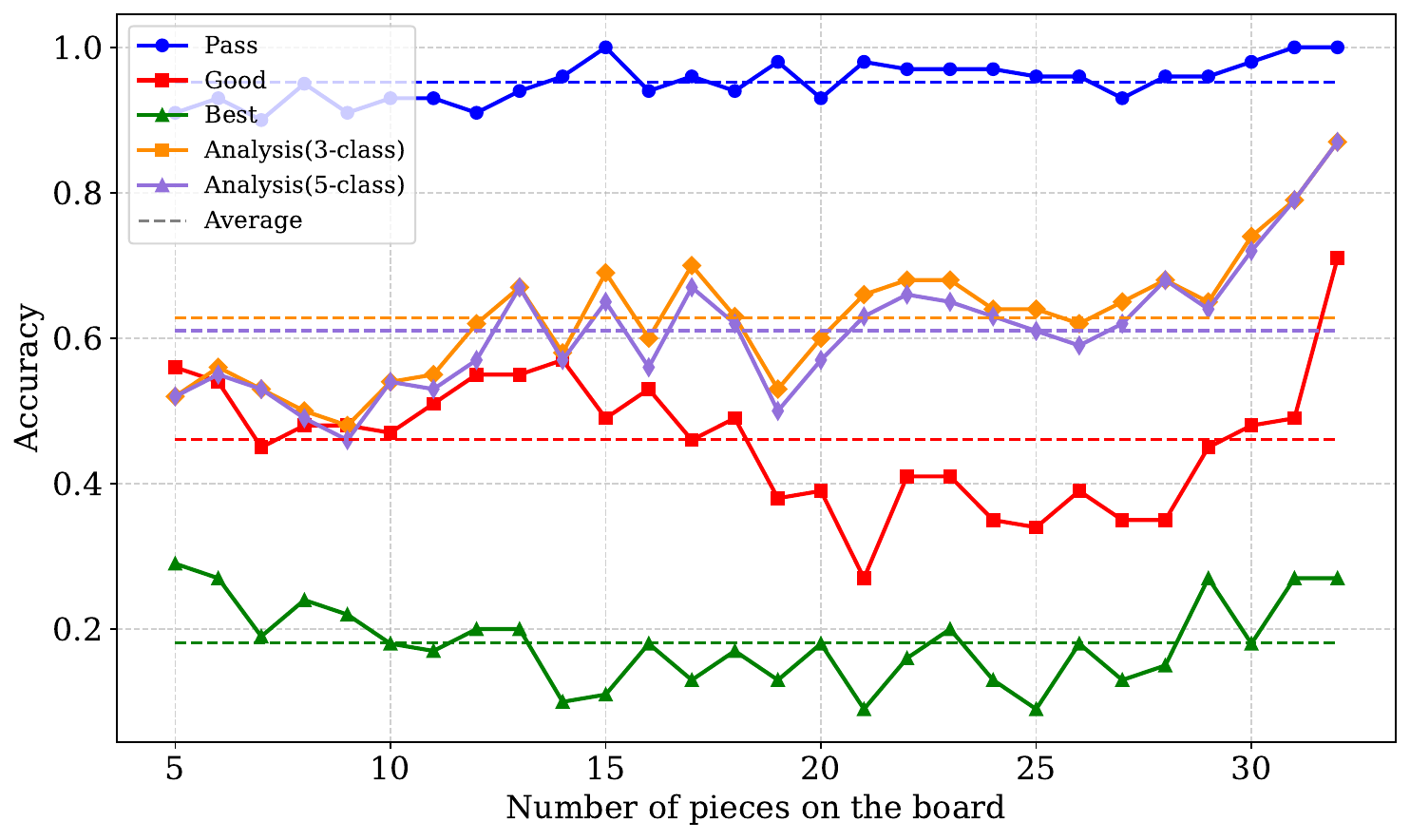}
        \caption{Performance of Xiangqi-R1 across piece counts.}
        \label{fig:move}
    \end{minipage}
    \hfill
    % --- 图2 ---
    \begin{minipage}[b]{0.3\textwidth}
        \centering
        \includegraphics[width=\textwidth]{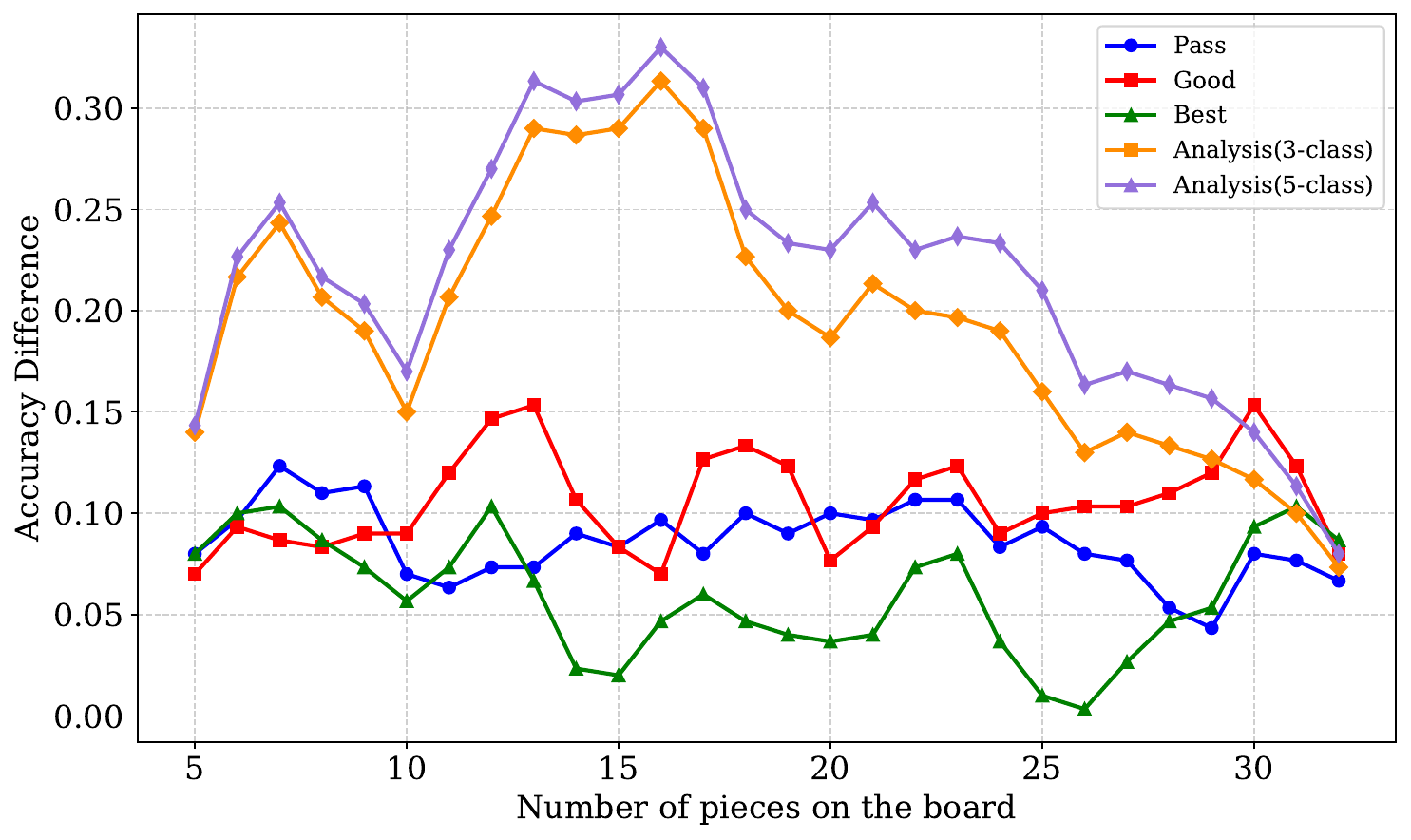}
        \caption{Xiangqi-R1 vs. Stage2: performance by piece count.}
        \label{fig:diff}
    \end{minipage}
    \hfill
    % --- 图3 ---
    \begin{minipage}[b]{0.3\textwidth}
        \centering
        \includegraphics[width=\textwidth]{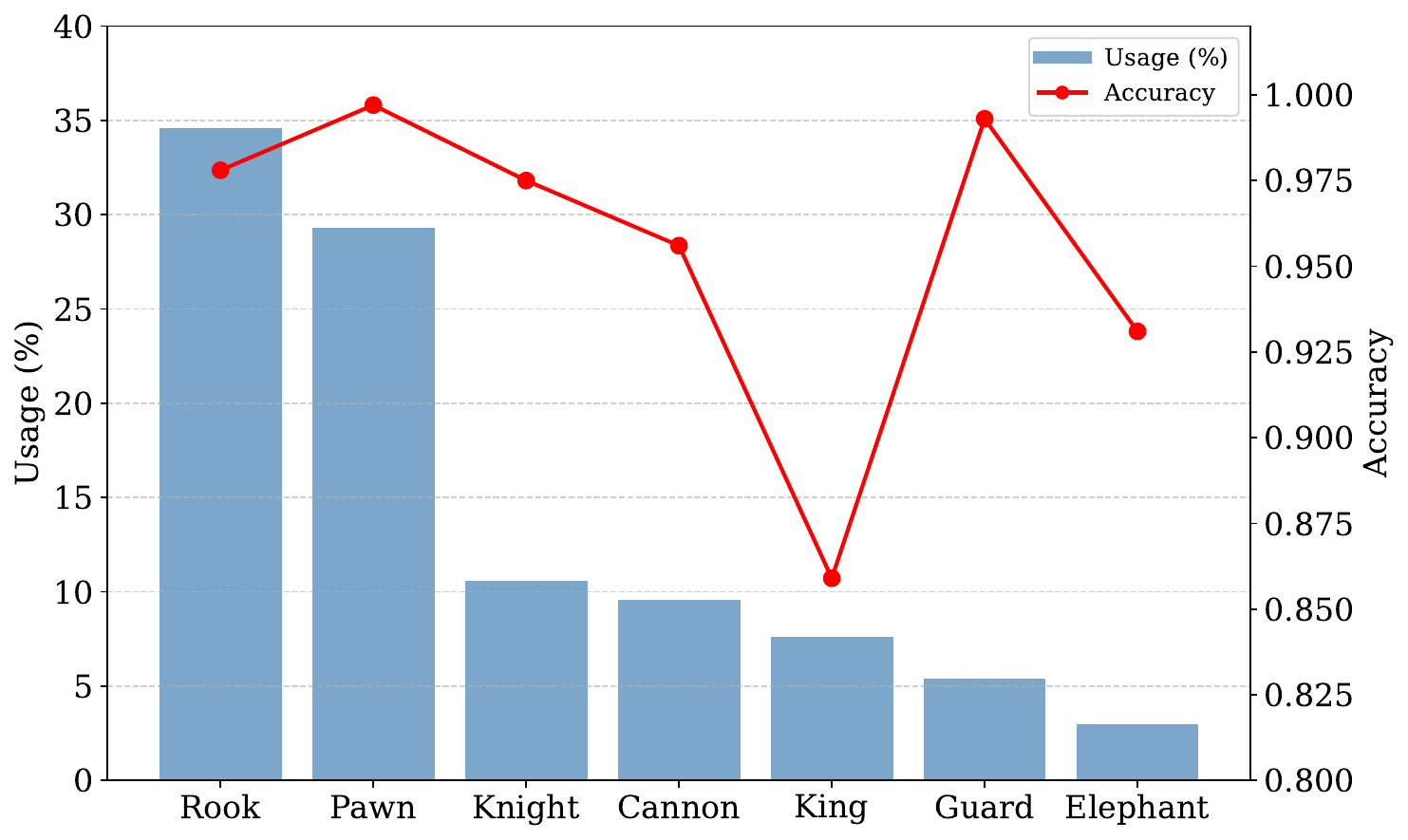}
        \caption{Xiangqi-R1 accuracy and usage by piece type.}
        \label{fig:temp}
    \end{minipage}

\end{figure*}

Tab.~\ref{tab:ablation} presents the results of our ablation study on the two core tasks. Compared with the vanilla model and the model fine-tuned only with supervised data (SFT-Stage2), our proposed Xiangqi-R1-7B achieves significantly better performance across all metrics, validating the effectiveness of our reward design. We further perform controlled ablations to isolate the impact of each reward component. When removing the move-related reward $R_\mathrm{move}$, the model's performance on the move suggestion task drops substantially, particularly on metrics like legal@1 (from 0.952 to 0.874) and best@1 (from 0.181 to 0.130). This demonstrates that $R_\mathrm{move}$ plays a critical role in helping the model generate legal and high-quality moves. On the other hand, removing the situation analysis reward $R_\mathrm{analysis}$ leads to a noticeable performance degradation in the situation analysis task. Specifically, the 3-class@1 and 5-class@1 accuracies drop by approximately 14.4\% and 16.6\%, respectively. This confirms that $R_\mathrm{analysis}$ is essential for enhancing the model's reasoning ability. The ablation results demonstrate that our reward design is not only effective in improving spatial strategic reasoning, but also that each reward term serves a distinct and synergistic function in training.

% \begin{figure}[t] 
% \centerline{
% \includegraphics[width=0.4\textwidth]{figs/move_full.pdf} } 
% \caption{ 
% Performance variation of Xiangqi-R1 across different numbers of pieces on the board.
% } 
% \label{fig:move} 
% \end{figure}
\subsection{Further analysis}
\paragraph{Our experiments show that piece count significantly affects model performance.} We evaluate the model on 2,800 test positions spanning 5–32 pieces. Fig.~\ref{fig:move} reports top-1 scores for three key metrics across these ranges. The legal@1 metric remains consistently high (around 0.95) with minimal variance, indicating that the model almost always generates legal moves. However, best@1 performance decreases notably in mid-range piece counts, roughly corresponding to the middlegame, likely due to increased complexity and strategic demands. A similar trend appears in good@1, with significant performance drops between 19 and 28 pieces. All three metrics reach their peak at high piece counts, typical of the opening.

In contrast, both situation-analysis metrics show a steady decline as piece count decreases. We attribute this to the rising difficulty of evaluating later-game positions, where smaller margins for error make each decision more consequential. Even so, the model performs strongly in the opening phase, where the game is more structured and predictable.

\paragraph{RL improves model performance across piece counts, with notable gains in situation analysis.} We compared ours model Xiangqi-R1 with the SFT-Stage2 model across different stages of the game, as shown in Fig.~\ref{fig:diff}. Our findings show that Xiangqi-R1 outperforms the SFT-Stage2 model on almost all metrics throughout the various game stages. Specifically, the legal@1 and good@1 metrics exhibit gains close to 10\% at each stage, demonstrating the effectiveness of our approach. In contrast, the best@1 metric shows a more limited improvement of around 5\% during the midgame phase. As discussed earlier, the complexity of the game state is highest in this stage, making it challenging for a 7B model. We believe that increasing the model size could further enhance reasoning performance during complex stages.

Remarkably, metrics related to situation analysis show gains of 15\%–30\%, underscoring the model’s strong potential in position evaluation, effectively enhanced by our reward design. However, it is worth noting that the improvements across all metrics during the opening phase are relatively modest, primarily because the SFT-Stage2 model already performs well in this stage.
% \begin{figure}[t] 
% \centerline{
% \includegraphics[width=0.4\textwidth]{figs/move_diff.pdf} } 
% \caption{ 
% Differences in performance between Xiangqi-R1 and SFT-Stage2 model across different board piece counts.
% } 
% \label{fig:diff} 
% \end{figure}
% 补充一下新的实验
\paragraph{Xiangqi-R1 shows significantly better proficiency with high-usage offensive pieces than with rule-bound defensive ones.} We systematically evaluate LLM performance in executing action commands for various Xiangqi pieces based on success rates and usage frequencies. The model performs well on frequently used pieces with simple movement rules, such as the \textit{Pawn} and \textit{Rook}, but its success rates decline for pieces with stricter or more context-dependent rules, including the \textit{King} and \textit{Elephant}.

In particular, the \textit{King} shows the lowest success rate (0.859). Although its movement appears simple, strict constraints such as avoiding direct confrontation with the opposing \textit{King} and restrictions when in check require precise recognition of the board state, and errors often arise from misinterpreting these rules. This underscores the complexity of the task. Nevertheless, our method substantially improves performance, achieving an overall success rate above 95\% across all piece types.

\section{Conclusion}
In this paper, we introduce Xiangqi-R1, a 7B-parameter model tailored for spatial strategic reasoning in Chinese Chess. Our multi-stage training framework integrates large-scale expert-annotated data, engine evaluations, and GRPO-based reinforcement learning with multi-dimensional rewards. Experiments show that Xiangqi-R1 significantly outperforms much larger general-purpose LLMs in move legality and strategic analysis. This work establishes a strong basis for advancing strategic intelligence in complex spatial domains.

\section{Limitations}

Due to computational resource constraints, our experiments were limited to 7B-parameter model. And we only employed text-based LLMs, which currently cannot process visual board inputs. In full-game experiments, we observed that neither our model nor general-purpose LLMs could reliably complete an entire game, highlighting the intrinsic difficulty of the task and the current model limitations. 
Nevertheless, our approach offers a promising direction for this field. Future work will focus on improving endgame reasoning, incorporating visual input, and enhancing full-game play capability.

\bibliography{custom}

\appendix
\section{Hyperparameters}\label{Hyperparameters}
All experiments were conducted on 4 * NVIDIA A6000 GPUs. The training process consists of three stages: (1) fine-tuning the model to predict legal moves, enabling it to learn fundamental spatial rules; (2) incorporating strategic annotations to enhance decision-making capabilities; and (3) applying RL with GRPO based on multi-dimensional reward signals to improve reasoning ability. Specifically, we initialized from Qwen-2.5-7B-Instruct and, across all three training stages, employed BF16 mixed-precision training, gradient checkpointing, and the DeepSpeed ZeRO-2 optimization strategy with CPU offloading to improve training efficiency and reduce GPU memory consumption. During the evaluation phase, we used the default inference settings and set \textit{max new tokens} to 2048. The specific parameter settings are detailed in Tab.~\ref{tab:hyperparameters}. Algorithm~\ref{alg:training} outlines the pseudocode for training the Xiangqi-R1.

\begin{table}[htbp]
    \centering
    \begin{tabular}{@{}lll@{}}
        \toprule
        \textbf{Stage} & \textbf{Parameter} & \textbf{Value} \\
        \midrule
        \multirow{6}{*}{Stage 1: SFT} 
        & learning rate & 1e-4 \\
        & batch Size & 64 \\
        & gradient acc steps & 4 \\
        & epochs & 1 \\
        & lora rank (\(r\)) & 16 \\
        & lora scaling (\(\alpha\)) & 32 \\
        & lora dropout & 0.1 \\
        \midrule
        \multirow{6}{*}{Stage 2: SFT} 
        & learning rate & 1e-4 \\
        & batch size & 32 \\
        & gradient acc steps & 2 \\
        & epochs & 3 \\
        & lora rank (\(r\)) & 16 \\
        & lora scaling (\(\alpha\)) & 32 \\
        & lora dropout & 0.1 \\
        \midrule
        \multirow{7}{*}{Stage 3: RL} 
        & learning rate & 1e-4 \\
        & batch size & 8 \\
        & gradient acc steps & 4 \\
        & epochs & 2 \\
        & lora rank (\(r\)) & 16 \\
        & lora scaling (\(\alpha\)) & 32 \\
        & lora dropout & 0.1 \\
        & rollout & 8 \\
        \bottomrule
    \end{tabular}
    \caption{Hyperparameter Settings for Training. }
    \label{tab:hyperparameters}
\end{table}

\begin{algorithm}
\caption{Three-Stage Xiangqi Model Training Pipeline}
\begin{algorithmic}[0]

\State Split dataset $\mathcal{D} \rightarrow \{\mathcal{D}_{\text{Stage1}}, \mathcal{D}_{\text{Stage2}}, \mathcal{D}_{\text{Stage3}}\}$

\Statex \textbf{Stage 1: SFT with Board-Move Data}
\For{each game state $d \in \mathcal{D}_{\text{Stage1}}$}
    \State Filter data $d$ by checking move legality and board validity
    \State Convert filtered data into (Board, Move) pairs
\EndFor
\State Train initial model $\pi_{\text{init}}$ on curated data $\mathcal{D}_{\text{Stage1}}$ via SFT to obtain policy $\pi_{\text{Stage1}}$

\Statex \textbf{Stage 2: SFT with Expert Commentary}
\For{each game state $d \in \mathcal{D}_{\text{Stage2}}$}
    \State Apply LLM-based rewriting to anonymize expert commentary
    \State Filter data using predefined quality and formatting rules
    \State Convert into (Board, Move, Comment) triplets
\EndFor
\State Train model $\pi_{\text{Stage1}}$ on verified data $\mathcal{D}_{\text{Stage2}}$ via SFT to obtain policy $\pi_{\text{Stage2}}$

\Statex \textbf{Stage 3: Reinforcement Learning with GRPO}
\For{each training iteration}
    \State Sample a batch of game states $d \sim \mathcal{D}_{\text{Stage3}}$
    \State Use $\pi_{\text{Stage2}}$ to generate move suggestions and corresponding analyses
    \State Compute total reward $R = R_{\text{move}} + R_{\text{analysis}} + R_{\text{format}}$
    \State Update model using GRPO: $\pi_{\text{Stage3}} \leftarrow \pi_{\text{Stage2}}$
\EndFor

\State Set final policy $\pi_{\text{final}} \gets \pi_{\text{Stage3}}$
\State \textbf{return} $\pi_{\text{final}}$ (Xiangqi-R1)

\end{algorithmic}
\end{algorithm}\label{alg:training}

\section{Xiangqi}\label{app:xiangqi}

Xiangqi, also known as Chinese Chess, is a traditional two-player, deterministic, fully observable, zero-sum board game. The game is played on a $9 \times 10$ grid whose geometry is distinguished by a central \textit{River} that divides the two sides and imposes movement constraints on specific piece types. Each player controls seven distinct types of pieces, each endowed with unique movement and capture mechanics. A $3 \times 3$ \textit{Palace} is located at the center of each player's back rank, restricting the movement of the \textit{King} and the \textit{Guards}. The objective of the game is to checkmate the opponent's King. The detailed movement rules of all piece types are summarized in Table~\ref{tab:xiangqi}.

\section{Data Collection}
To efficiently train and evaluate Xiangqi-R1, we constructed a high-quality, standardized Xiangqi dataset. Given the large scale of the data, we plan to release anonymized dataset publicly in a manner that is appropriate. The dataset primarily originates from several publicly accessible professional Xiangqi game record platforms, which provide extensive match logs from high-level players. Below, we describe the key strategies and procedures employed during data collection, cleaning, transformation, and filtering.

\subsection{Data Sources and Initial Filtering}
We collected a total of \textbf{182,342} game records, all standardized in the Xiangqi Portable Game Notation (PGN) format. Each record contains the complete game process, move sequences, game results, and, in some cases, expert commentary and annotations. To ensure data integrity and usability, we excluded records with erroneous move sequences, unparseable moves, or missing result labels (win, loss, or draw). After this initial cleaning, we retained a high-quality subset of PGN data for subsequent processing.

\subsection{Data Structure Transformation}
To facilitate model training and engine analysis, the original PGN data were uniformly converted to Forsyth–Edwards Notation (FEN) format. Each move was represented as a triplet (FEN, Move, Comment). This transformation was conducted using the open-source tool \texttt{cchess}, which automatically parses moves and generates position encodings. During the conversion, we performed the following structured processing:
\begin{itemize}[noitemsep, topsep=0pt, leftmargin=1em]
\item Generated the current position’s FEN and the corresponding move for each turn;
\item Retained expert commentary in the Comment field when available, otherwise left it empty;
\item Treated each record as an independent sample to facilitate state-specific predictive learning.
\end{itemize}

Additionally, we preserved only the moves made by the winning side in decisive games, as well as all moves from both sides in drawn games. This approach is based on the observation that although the winning side’s moves may contain errors, their overall move logic tends to be reasonable, and mistakes are often corrected in later stages. Conversely, the losing side’s errors tend to be decisive. This filtering resulted in \textbf{6,919,633} samples.

\subsection{Data Processing and Partitioning}

To protect user and player privacy, we applied the LLMs DeepSeek-V3 to sanitize commentary content by automatically detecting and removing sensitive information such as real names, locations, and event details.

Furthermore, we excluded comments that did not contain any Xiangqi-related keywords (e.g., “red,” “black,” “rook,” “knight,” “cannon,” etc.). Only expert comments related to \emph{high-quality moves} were retained. We defined GoodMove as those whose evaluation value NewValue, computed by the Pikafish engine (set to depth 25 in our experiments), is within 100 centipawns of the best move’s evaluation BestValue under given parameters, as formalized in Eq.~\ref{eq:good}.

Following these preprocessing steps, we partitioned the dataset into three training stages and one independent test set. To support position evaluation tasks, we used Pikafish evaluation scores to label positions in Stage 2 and Stage 3. Additionally, we constructed a three-class labeling scheme by merging the “slight advantage” and “clear advantage” categories into a unified “advantage” class, as formalized in Eq.~\ref{eq:situation_analysis}. The test set consists of 2,800 carefully selected high-quality samples, with 50 positions each for red and black sides, spanning piece counts between 5 and 32.

The final data partitioning is summarized as follows:
\begin{itemize}[noitemsep, topsep=0pt, leftmargin=1em]
\item \textbf{Stage 1:} 5,109,543 (FEN, Move) samples for basic move prediction training;
\item \textbf{Stage 2:} 52,913 (FEN, Move, Comment) annotated samples for situation analysis and decision-making;
\item \textbf{Stage 3:} 52,800 (FEN, Move, Comment, Score) annotated samples focusing on enhancing
reasoning stability;
\item \textbf{Test Set:} 2,800 high-quality samples covering opening, midgame, and endgame phases, for model evaluation.
\end{itemize}

It is worth noting that in Stage 3 and the test set, all legal moves for each FEN position were exhaustively enumerated, and their evaluation scores computed via Pikafish to facilitate subsequent evaluation, analysis, and classification tasks.

\section{Prompt}
In this section, we present the prompts used in our experiments, which were originally written in Chinese. For ease of reading, we provide their English translations. It is worth noting that during Stage 1 of training, we omit the Situation Analysis section in Fig.~\ref{fig:prompt_train}, as the model is only required to generate what it considers the best move at this stage.

%filter
% Please revise the following Xiangqi (Chinese Chess) commentary sentences. Without changing the original meaning or key technical terms, improve the sentence structure and punctuation for clarity and fluency. Additionally, remove any irrelevant information (e.g., personal background or unrelated context). 
% Output format:{'comment': 'Revised sentence'}

\section{Baseline}
Our results demonstrate that Xiangqi-R1 consistently outperforms open-source models of comparable or larger scales. Among these baseline models, some were accessed via API calls while others were locally deployed. All API-based models were accessed through the Volcano Engine platform, using official endpoints provided by the respective model developers. The specific version numbers of all models used in our experiments are listed below.

\begin{table}[ht]
  \centering
  \small
  \resizebox{0.5 \textwidth}{!}{
  \begin{tabular}{l|c|l}
    \toprule
    \textbf{Model} & \textbf{Size} & \textbf{Version (or Hugging Face ID)} \\
    \midrule
    \multicolumn{3}{c}{\textbf{API Access}} \\
    \midrule
    R1-Distill-Qwen-7B & 7B & r1-distill-qwen-7b-250120 \\
    Doubao-Lite & - & doubao-1-5-lite-32k-250115\\
    Doubao-Pro-1.5 & - & doubao-1-5-pro-32k-250115 \\
    Doubao-Pro-1.5-Think & - & doubao-1-5-thinking-pro-250415 \\
    DeepSeek-V3 & 671B & deepseek-v3-250324 \\
    DeepSeek-R1 & 671B & deepseek-r1-250528 \\
    
    \midrule
    \multicolumn{3}{c}{\textbf{Local Deployment}} \\
    \midrule
    Qwen-2.5-7B-Instruct & 7B & Qwen/Qwen2.5-7B-Instruct \\
    Xiangqi-R1-7B (Ours) & 7B & - \\
    \bottomrule
  \end{tabular}
  }
  \caption{Model list used in our evaluation, grouped by access method. }
  \label{tab:model_list}
\end{table}

\begin{table*}[htbp]
\centering
\resizebox{\textwidth}{!}{
\begin{tabular}{@{}ll@{}}
\toprule
\textbf{Piece} & \textbf{Movement Rule} \\ \midrule
\textit{King} & Moves one point orthogonally within the \textit{Palace}; must not be placed in direct line of sight with the opposing King. \\
\textit{Rook} & Moves any number of unobstructed points orthogonally. \\
\textit{Knight} & Moves one orthogonal step followed by one diagonal step; the move is invalid if the orthogonal ``leg'' is blocked. \\
\textit{Cannon} & Moves like a Rook in non-capturing moves; captures by leaping over exactly one intervening piece. \\
\textit{Elephant} & Moves two points diagonally; cannot cross the River and is blocked if the midpoint is occupied. \\
\textit{Guard} & Moves one point diagonally and must remain inside the Palace. \\
\textit{Pawn} & Moves one point forward; after crossing the River, it may also move one point horizontally. \\
\bottomrule
\end{tabular}
}
\caption{Movement rules of Xiangqi pieces.}
\label{tab:xiangqi}
\end{table*}

\begin{figure}[htbp]
    \centering
\begin{tcolorbox}[colback=white,colframe=black, enhanced jigsaw, listing only, listing options={basicstyle=\rmfamily}]

\textbf{Task Description:} \\
Please revise the following Xiangqi (Chinese Chess) commentary sentences. Without changing the original meaning or key technical terms, improve the sentence structure and punctuation for clarity and fluency. Additionally, remove any irrelevant information (e.g., personal background or unrelated context). 
\\
\textbf{Output Format:} \\
\{'comment': 'Revised sentence'\}

\end{tcolorbox}
\caption{Prompts used to filter out irrelevant content.}
\label{fig:prompt_filter}
\end{figure}

\begin{figure*}[htbp]
\centering
\begin{tcolorbox}[colback=white,colframe=black, enhanced jigsaw, listing only, listing options={basicstyle=\rmfamily}]

You are a Xiangqi (Chinese Chess) master. Below is the current game position. Please analyze the situation and recommend the best next move.

\textbf{Task Requirements}:\\
- Evaluate the current position using one of the following: balanced, slight advantage for Red, slight advantage for Black, significant advantage for Red, or significant advantage for Black.\\
- Recommend your best next move.

\textbf{Xiangqi Rules (Summary)}:\\
- Xiangqi is a two-player strategy game with 16 pieces per side. Players take turns, and the goal is to checkmate the opponent.\\
- K/k (King): moves one step vertically or horizontally within the palace; cannot face the opposing king directly across the board.\\
  - A/a (Guard): moves one step diagonally within the palace.\\
  - B/b (Elephant): moves two steps diagonally, cannot cross the river, and cannot move if the midpoint (elephant eye) is occupied.\\
  - N/n (Knight): moves in an L-shape (one step straight, then one step diagonal); cannot move if the straight path is blocked.\\
  - R/r (Rook): moves any number of steps along a straight line.\\
  - C/c (Cannon): moves like a chariot but captures only by jumping over exactly one piece.\\
  - P/p (Pawn): moves one step forward before crossing the river, and may move sideways after crossing, but never backward.

\textbf{Move Format (ICCS)}:\\
- ICCS (Internet Chinese Chess Server) notation uses a four-character format: starting position + ending position.\\
- Columns (files) are labeled a–i (left to right), and rows (ranks) are 0–9 (bottom to top).\\
- Example: `h2e2` means moving the piece from square h2 to e2.

\textbf{FEN Explanation}:\\
- The board layout is represented in Forsyth–Edwards Notation.\\
- Lowercase letters = Black pieces; uppercase = Red.\\
- Piece codes: K/k (King), A/a (Guard), B/b (Elephant), N/n (Knight), R/r (Rook), C/c (Cannon), P/p (Pawn).\\
- Numbers represent consecutive empty squares; slashes separate the 10 ranks.\\
- `w` means Red to move; `b` means Black to move.

\textbf{Board FEN}: \textcolor{gray}{\texttt{<FEN>}}

\textbf{Text Description of Board}: \textcolor{gray}{\texttt{<Board Text>}}

\textbf{Model Answer}:
Please return your analysis and best move using the following format:\\
Situation Analysis: xxxx  \\
Best Move: xxxx

\end{tcolorbox}
\caption{Prompts designed for model training and evaluation.}
\label{fig:prompt_train}
\end{figure*}

\end{document}